\documentclass[12pt]{iopart}

\pdfoutput=1
\usepackage[utf8]{inputenc}
\usepackage[USenglish]{babel}

\usepackage{fancyhdr}
\usepackage{microtype}
\usepackage{graphicx}
\usepackage{subfigure}
\usepackage{booktabs}
\usepackage{amssymb,amsthm,amsfonts}
\usepackage{MnSymbol}
\usepackage{mathrsfs}
\usepackage{slashed}
\usepackage{color,rotating}
\usepackage{dsfont}
\usepackage{setspace}
\usepackage{physics}
\usepackage{verbatim}
\usepackage{hyperref,doi}
\usepackage[export]{adjustbox}
\usepackage[normalem]{ulem}
\usepackage{bbold}
\usepackage[linesnumbered,ruled,vlined]{algorithm2e}
\SetKwInput{KwInput}{Input}    
\SetKwInput{KwOutput}{Output}

\usepackage{graphicx}

\usepackage[labelfont=bf, font=footnotesize]{caption}

\makeatletter
\def\l@subsubsection#1#2{}
\makeatother

\newcommand{\beq}{\begin{equation}}
\newcommand{\eeq}{\end{equation}}
\newcommand{\beqa}{\begin{eqnarray}}
\newcommand{\eeqa}{\end{eqnarray}}
\newcommand{\bfc}{\begin{figure}[t]\begin{center}}
\newcommand{\efc}{\end{center}\end{figure}}

\def\Fig#1{Fig.~\ref{#1}}
\def\fig#1{Fig.~\ref{#1}}
\def\tab#1{Table~\ref{#1}}

\def\eq#1{(\ref{#1})}

\def\sec#1{Section~\ref{#1}}
\def\0#1#2{\frac{#1}{#2}}  

\graphicspath{{./figures/}}

\newcommand{\be}{\begin{eqnarray}}
\newcommand{\ee}{\end{eqnarray}}

\begin{document}

\title[Twin Neural Network Regression Algorithm]{Twin Neural Network Improved k-Nearest Neighbor Regression}

\author{Sebastian J. Wetzel}
\address{University of Waterloo, Waterloo, Ontario N2L 3G1, Canada}
\address{Perimeter Institute for Theoretical Physics, Waterloo, Ontario N2L 2Y5, Canada}
\address{Homes Plus Magazine Inc., Waterloo, Ontario N2V 2B1, Canada}
\ead{swetzel@perimeterinstitute.ca}


\begin{abstract}
Twin neural network regression is trained to predict differences between regression targets rather than the targets themselves. A solution to the original regression problem can be obtained by ensembling predicted differences between the targets of an unknown data point and multiple known anchor data points. Choosing the anchors to be the nearest neighbors of the unknown data point leads to a neural network-based improvement of k-nearest neighbor regression. This algorithm is shown to outperform both neural networks and k-nearest neighbor regression on small to medium-sized data sets. 
\end{abstract}
\vspace{2pc}
\maketitle


\newpage
\section{Introduction}

Regression is a computational method used to predict numerical values based on input data. It focuses on understanding and quantifying relationships between variables, allowing for informed decision-making and forecasting future outcomes. This technique is versatile and can handle various types of data, ranging from simple linear relationships to more complex, non-linear interactions. Regression models learn from historical data patterns and use this knowledge to make accurate predictions on new, unseen data.

There are several machine learning models that can be used for regression tasks, each with its own strengths and suitability for different types of data and problem domains. Besides others, these include K-Nearest Neighbors (k-NN) Regression. This method predicts the target value for a given data point by averaging the values of its k-nearest neighbors. For the purpose of this article, of note is Neural Network Regression which is capable of capturing complex relationships in the data.

Twin neural network regression (TNNR) is a machine learning technique based on artificial neural networks. It is a specialized approach that aims to predict the difference between target values rather than directly predicting the target values themselves. The original regression problem is solved by aggregating the predicted differences between the target values of an unknown data point and several known anchor data points. This ensemble approach leverages the collective information from multiple sources to arrive at a solution for the regression problem. The methodology behind TNNR offers several advantages \cite{wetzel2020twin,wetzel2021twin,tynes2021pairwise}:
\begin{itemize}
\item{Ensemble Predictions:} By generating a single prediction for each anchor data point, TNNR effectively creates an ensemble of predictions from a single trained model.

\item{Improved Accuracy:} TNNR can be particularly effective in scenarios where accurate predictions are needed, especially when the available training data is limited.

\item{Uncertainty Estimation:} TNNR can provide valuable uncertainty estimates associated with its predictions, which can be crucial in decision-making processes.

\item{Semi-Supervised Learning:} TNNR can be extended to facilitate semi-supervised learning, where it enforces consistency conditions on unknown data points through a modified loss function.

\item{Beneficial in Data-Scarce Domains:} TNNR tends to perform well in domains where obtaining sufficient training data is challenging or costly. However, it scales poorly on large data sets.
\end{itemize}

In this manuscript I aim to combine TNNR with k-Nearest Neighbors (k-NN) regression which predicts a continuous value for a data point by averaging the values of its nearest neighbors in the dataset. Its advantages are the following:

\begin{itemize}
\item{Simplicity:} k-NN regression is straightforward and easy to understand, making it a good choice for beginners in machine learning.

\item{Non-Parametric:} It doesn't make any assumptions about the underlying data distribution, which can be beneficial in situations where the data may not follow a specific mathematical model.

\item{No Training Phase:} Unlike many other algorithms, k-NN doesn't require a training phase. It simply stores the dataset, making it efficient for quick implementation.

\item{Effective for Small Datasets:} k-NN can be particularly effective when the dataset is small, as it doesn't require a large amount of training data to make accurate predictions. However, computationally inefficient for large datasets, and has difficulties handling high-dimensional data. 
\end{itemize}

The goal of this article is to combine k-NN regression and TNNR. One the one hand this can be viewed as choosing a subset of nearest neighbors as anchor data points for TNNR. On the other hand, it might also be viewed as training an artificial neural network to predict the magnitude of adjustments that need to be applied to the value of a certain nearest neighbor before averaging.

\subsection{Outline}

This article is structured in the following manner: After explaining prior research on k-nearest neighbors and twinned regression methods also known as pairwise difference regression methods in \sec{chapter:Prior Work}:Prior Work, I describe the mathematical formulation in \sec{chapter:Review of Twin Neural Network Regression}: Reformulation of the Regression Problem. Next, I describe the algorithm to combine k-NN regression and TNNR in \sec{chapter:Nearest Neighbor TNNR}:Nearest Neighbor Twin Neural Network Regression. Finally, I present the results in \sec{chapter:results}:Results, examine the time scaling in \sec{chapter:Time Scaling}: Time Scaling and and draw conclusions in \sec{chapter:conclusion}: Conclusion.

\section{Prior Work}
\label{chapter:Prior Work}

The purpose of this article is to explain how to combine the k-nearest neighbors algorithm \cite{cover1967nearest,fix1989discriminatory} with twin neural network regression \cite{wetzel2020twin} to obtain accurate predictions for a regression problem. Going forward I call this algorithm: Nearest neighbor twin neural network regression (NNTNNR).

k-NN regression builds upon the k-nearest neighbors algorithm 
\cite{cover1967nearest,fix1989discriminatory} used to find the nearest neighbors of an unlabelled data point. The regression prediction is calculated by a (weighted) average of the values of k-nearest neighbors. Standard k-NN has been improved by a clever choice of important neighbors and smart weighting of each neighbors contributions \cite{wang2007improving,gou2011novel,kuhkan2016method,ertuugrul2017novel,pan2017new,song2017efficient,syaliman2018improving}. An improved version of k-NN can be learned by graph neural networks \cite{kang2021k}. Further, artificial neural networks have been employed in tandem with k-NN regression in different contexts before \cite{wu2009novel,bensaci2021deep,liu2018hybrid}.

The pairwise comparison inherent to twinned regression methods is inspired by Siamese neural networks which were devised to solve the similarity classification problem as it occurs in fingerprint recognition or signature verification \cite{bromley1993signature,baldi1993neural}. Siamese neural networks contain two identical neural networks with shared weights that project a pair of inputs into a latent space on which the pairwise similarity is determined by the distance. Twinned regression methods also take a pair of inputs to predict the difference between the regression targets \cite{wetzel2020twin}.

Twin neural network regression \cite{wetzel2020twin} was invented as a regression method that is trained to predict pairwise differences between the target values of pairs of input data points. This can be used to obtain more accurate predictions of differences than solving the original regression problem and then manually calculating the difference \cite{fralish2023deepdelta}. More importantly, the solution to the original regression problem is then obtained by aggregating the predicted differences between an unknown data point and several known anchor data points. Later, the same idea has been independently developed for random forests \cite{tynes2021pairwise}. This kind of regression framework has been shown to have several advantages: (1) it allows for a very efficient generation of ensemble predictions \cite{wetzel2020twin,tynes2021pairwise}. Typically, in methods that generate ensembles from training a single machine learning model, the predictions are strongly correlated \cite{srivastava2014dropout,wan2013regularization} since they can be deformed into each other through small perturbations. In twinned regression methods, however, ensemble members are separated by the distance of the input data points themselves. (2) Twinned regression methods tend to be more accurate than the underlying base algorithm on many data sets \cite{wetzel2020twin,tynes2021pairwise}, (3) consistency conditions allow for the formulation of uncertainty estimators in addition to the ensemble variance \cite{wetzel2020twin,tynes2021pairwise,fralish2023deepdelta} and (4) loops containing unlabelled data points can be supplied while training, hence turning the method into a semi-supervised regression algorithm \cite{wetzel2021twin}. Further, (5) the intrinsic uncertainty estimation lends itself for active learning \cite{tynes2021pairwise}.

The data sets, see \ref{sec:data}, are aligned with the experiments in \cite{wetzel2020twin,wetzel2021twin} and are chosen such that they contain a range of standard regression problems of small to medium size in addition to scientific data sets that can be modeled by equations. Extensive hyperparameter testing for neural networks on these data sets has been performed in \cite{wetzel2020twin}.

\begin{figure*}[h!]
    \centering
    \includegraphics[width=0.9\textwidth]{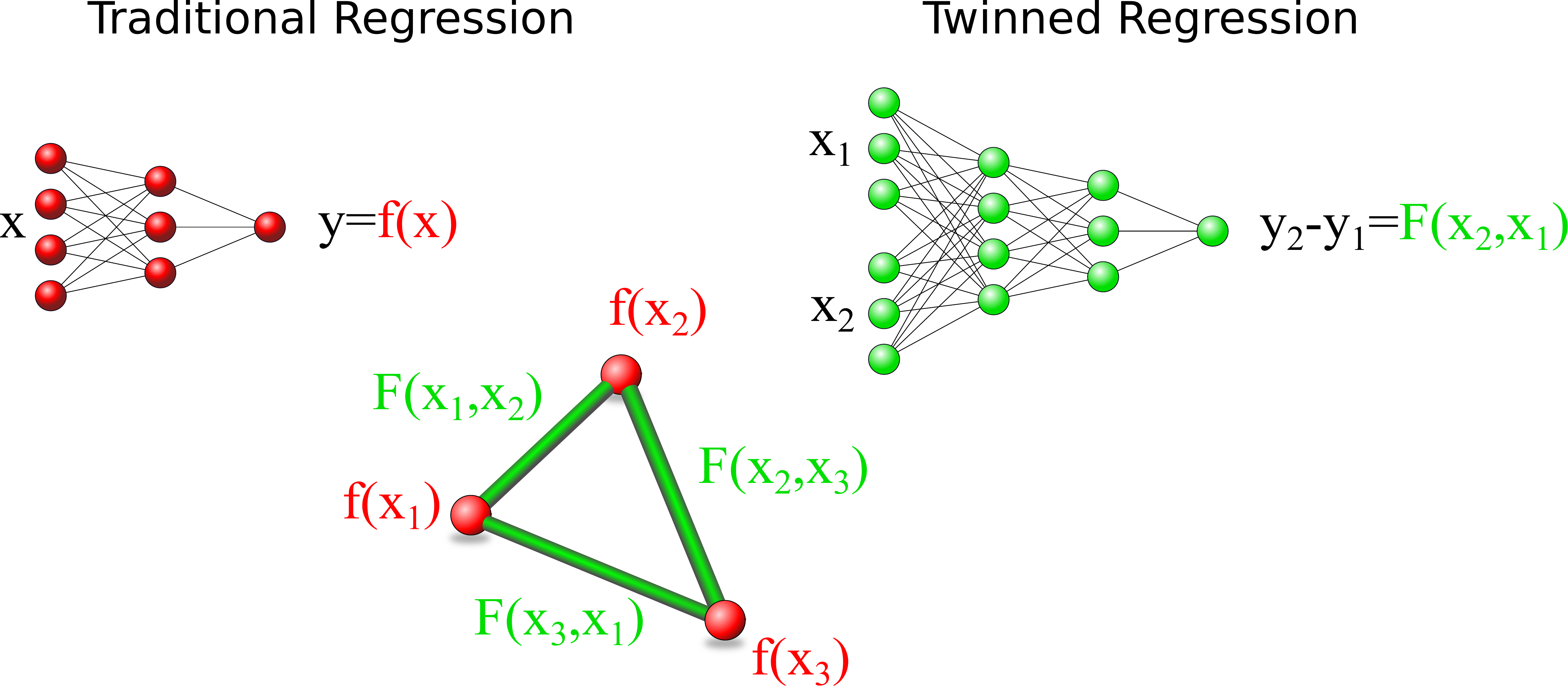}
    \caption{Reformulation of a regression problem: A traditional solution to a regression problem consists of finding an approximation to the function that maps a data point $x$ to its target value $f(x)=y$. Twinned regression methods solve the problem of mapping a pair of inputs $x_1$ and $x_2$ to the difference between the target values $F(x_2,x_1)=y_2-y_1$. The resulting function can then be employed as an estimator for the original regression problem $y_2=F(x_2,x_1)+y_1$ given a labelled anchor point $(x_1,y_1)$. Twinned regression methods must satisfy loop consistency: predictions along each loop sum to zero:  $F(x_1,x_2)+F(x_2,x_3)+F(x_3,x_1)=0$.}
    \label{fig:architecture}
\end{figure*}

\section{Review of Twin Neural Network Regression}
\label{chapter:Review of Twin Neural Network Regression}

A regression problem can be formulated as follows: Given a labelled training data set of $n$ data points $X^{train}=(x_1^{train},...,x_n^{train})$ with their corresponding target values $Y^{train}=(y_1^{train},...,y_n^{train})$ and a test data set $X^{test}=(x_1^{test},...,x_m^{test})$ of size $m$, we are tasked to find a function $f$ such that the deviation between $f(x_i)$ and $y_i$ is minimized with respect to a predefined objective function for all data points $x_i\in X^{train}\cup X^{test}$. In this work, this function is the root mean square error $L_{RMSE}=\sqrt{\sum_{i=1}^n (f(x_i)- y_i)^2}$. Unless stated otherwise, all performance measures are evaluated on unknown test data $(X^{test},Y^{test})$.

Twinned regression methods aim to solve a reformulation of the original regression problem which is visualized in \fig{fig:architecture}. For each pair of data points $(x_i^{train},x_j^{train})$ I train a regression model to find a function $F$ to predict the difference 
\begin{align}
    F(x_i,x_j)= y_i-y_j \quad .    \label{eq:TNNR}
\end{align}

This function $F$ can be used to construct a solution to the original regression problem via $y_i^{pred}= F(x_i,x_j)+y_j$, where $(x_j,y_j)$ is an {\em anchor} whose target value is known. Every training data point $x_j^{train}\in X^{train}$ can be used as such an anchor. A more accurate estimate for the solution of the original regression problem is obtained by averaging over many differences between a fixed unknown data point and different anchor data points
\begin{align}
    y_i^{pred}&=  \frac{1}{n}\sum_{j=1}^n \left(F(x_i,x_j^{train})+y_j^{train}\right)\nonumber \\
    &= \frac{1}{n}\sum_{j=1}^n\left( \frac{1}{2}F(x_i,x_j^{train})-\frac{1}{2}F(x_j^{train},x_i)+y_j^{train} \right) \quad .\label{eq01}
\end{align}
The increase in accuracy is based on averaging out the noise from different anchors and the reduction of the variance error via an ensemble of predictions. Previous works \cite{wetzel2020twin,tynes2021pairwise} recommended using the whole training data set as anchors, hence creating an ensemble of difference predictions $y_i-y_j$ which is twice as large as the training set for every single prediction of $y_i$. 

A major advantage of the dual formulation is the description via loops containing multiple data points as can be seen in \fig{fig:architecture}. In contrast to traditional regression, the results of twinned regression methods need to satisfy consistency conditions, for example for every three data points $x_1,x_2,x_3$, summing up the predictions along a closed loop should yield zero: $F(x_1,x_2)+F(x_2,x_3)+F(x_3,x_1)=0$. During inference, violations of these consistency conditions give rise to uncertainty 
estimates \cite{wetzel2020twin,tynes2021pairwise}. Enforcing loop consistency on predictions involving unlabelled data points in the training phase is what turns twinned regression methods into semi-supervised regression algorithms \cite{wetzel2021twin}.

\section{Nearest Neighbor Twin Neural Network Regression }
\label{chapter:Nearest Neighbor TNNR}

\begin{figure*}[h!]
    \centering
    \includegraphics[width=0.9\textwidth]{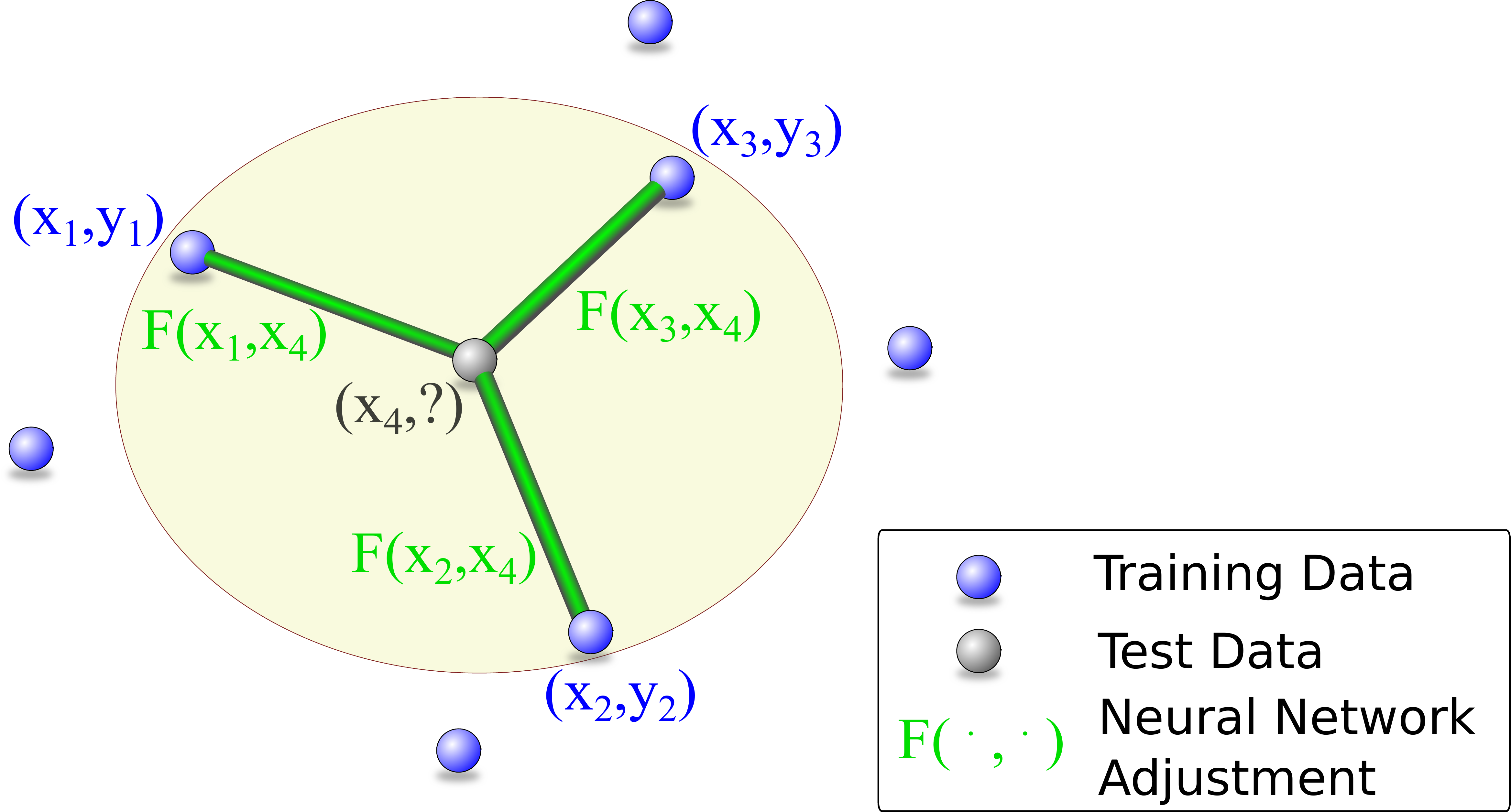}
    \caption{Nearest Neighbor Twin Neural Network Regression: An algorithm that improves the predictions of k-nearest neighbor regression with artificial neural networks. A k-nearest neighbor algorithm selects k-nearest neighbors of an unknown data point. A neural Network F calculates an adjustment to the $y$-values of the training data points. After adjusting, these values are averaged to obtain the final prediction.}
    \label{fig:NNTNNR}
\end{figure*}

\begin{algorithm}
\caption{Nearest Neighbor Twin Neural Network Regression Training}\label{alg:NNTNNRt}
  \KwData{Labelled data set $D=(X_{train},Y_{train})$} 
  \KwInput{Number of nearest neighbors  $k \in \mathbb{N}\cup \infty$}

  \SetKwFunction{FMain}{NN}
  \SetKwProg{Fn}{Function}{:}{}
  \Fn{\FMain{$x$, $X_{ref}$, $k$}}{
            $L$ $\leftarrow$ select k nearest neighbors of $x$ within $X_{ref}$\\
        \KwRet $L$
  }

  $\tilde{D}\leftarrow[((x_i,x_j),y_{ij}=y_i-y_j) $
  for $ x_j \in \FMain(x_i,X_{train},k)$
  for $ x_i \in X_{train}  ]$\\
  
  initialize twin neural network regression model $F$\\
  train $F$ on $\tilde{D}$\\

  \KwOutput{Trained model $F$}  
\end{algorithm}

\begin{algorithm}
\caption{Nearest Neighbor Twin Neural Network Regression Inference}\label{alg:NNTNNRi}
  \KwData{ Labelled data set $D=(X_{train},Y_{train})$\\ \phantom{\textbf{Data: }} Unlabelled data point $x_{test}$} 
  \KwInput{Trained twin neural network regression model $F$\\
  \phantom{\textbf{Input: }}Number of nearest neighbors  $k \in \mathbb{N}\cup \infty$}

  \SetKwFunction{FMain}{NN}
  \SetKwProg{Fn}{Function}{:}{}
  \Fn{\FMain{$x$, $X_{ref}$, $k$}}{
            $L$ $\leftarrow$ select k nearest neighbors of $x$ within $X_{ref}$\\
        \KwRet $L$
  }

  $\tilde{D}=(\tilde{X}_{inference},\tilde{Y}_{inference})\leftarrow[((x_{test},x_j),y_j) $
  for $ x_j \in \FMain(x_{test},X_{train},k)]$\\
  
  $P\leftarrow \text{mean}( F(X_{inference})+Y_{inference}$)

  \KwOutput{Prediction $P$}   		
\end{algorithm}

In this manuscript, I propose a new regression algorithm based on a combination of k-nearest neighbor regression and Twin Neural Network Regression called Nearest Neighbor Twin Neural Network Regression (NNTNNR) as depicted in \fig{fig:NNTNNR}. In principle, it could be implemented for various baseline twinned regression algorithms. In standard twinned regression methods the model learns to predict differences between the targets of two arbitrary data points. This model is then employed to create an ensemble prediction via averaging the approximations of the differences between the target value of a new data point and all anchor data points, see \eq{eq01}. However, not all of these anchor data points might be of equal importance for the prediction. That is why in this section I restrict the anchor points to the nearest neighbors. For this purpose, I define the notation $\text{NN}(i,m)$ as the set of $m$ nearest neighbors of a data point $x_i \in X$ within the training set $x_j \in X^{train}$ to reformulate the prediction:

\begin{align}
    y_i^{pred}&=  \frac{1}{m}\sum_{j\in \text{NN}(i,m)} \left( F(x_i,x_j^{train})+y_j^{train} \right)
\end{align}

I have defined the prediction using nearest neighbors during the inference phase. However, I still need to determine whether it is better to train the model to predict differences between target values of generic data points or between neighboring data points.

Trivially, NNTNNR can be related to k-NN regression through setting $F(x_i,x_j)\equiv 0$, then 

\begin{align}
    y_i^{pred}&=  \frac{1}{m}\sum_{j\in NN(i,m)} \left( \underbrace{F(x_i,x_j^{train})}_{0}+y_j^{train} \right)=\frac{1}{m}\sum_{j\in \text{NN}(i)}^m y_j^{train}
    \label{eq:nntnnr_knn}
\end{align}
On the one hand, assuming $F(x_i,x_j)$ would just be a minor contribution to k-NN regression one would see a qualitatively similar performance of NNTNNR. On the other hand, assuming the nearest neighbor anchor selection would just create an improved ensemble of predictions, the NNTNNR performance would be similar to standard TNNR.

The algorithm \ref{alg:NNTNNRt} for training this model is similar to training standard TNNR. However, one needs to make a choice whether one restricts the paired training set to only contain $k$ nearest neighbors pairs or whether to include the full paired training set which is equivalent to $k=\infty$. Similarly, the algorithm for the prediction on new unseen data points can be found in algorithm \ref{alg:NNTNNRi}. 

\section{Results}
\label{chapter:results}
In this section, I apply NNTNNR to 9 different data sets and compare the performance to traditional TNNR and k-NN regression. In the following experiments, I further differentiate between training on all possible training pairs and only nearest neighbor training pairs. All the results are collected in \tab{tab:nn_tnnr} and the performance while varying the number of nearest neighbors are visualized in \fig{fig:B} and \fig{fig:D}.

\subsection{Notes About Experiments}
All experiments in this article are performed on the data sets outlined in \ref{sec:data}. They use the full data sets of which 70\% are used for training, 10\% as validation set and 20\% as test set. The details of the neural network architectures can be found in the appendix \ref{sec:nn_architecture}. All experiments are repeated for 25 random but fixed splits of training, test, and if applicable, validation data.

\subsection{NNTNNR vs TNNR}
\begin{figure}
    \centering
    \includegraphics[width=0.95\textwidth]{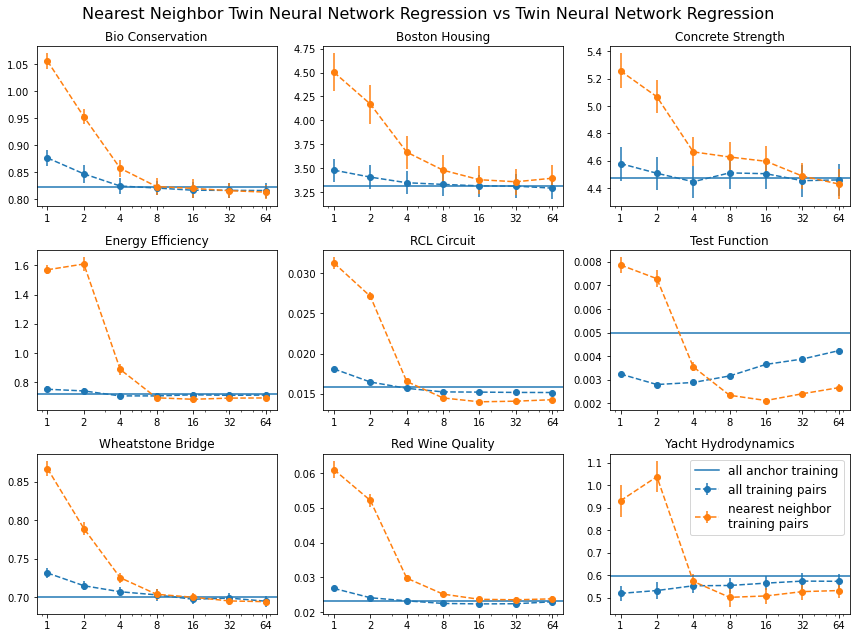}
    \caption{NNTNNR performance measured in terms of RMSE vs the number of nearest neighbors used if applicable. The solid blue line marks the performance of the original TNNR. The dashed blue line displays the results of NNTNNR trained on all possible pairs while performing inference only using nearest neighbor anchors. The dashed orange line is produced by restricting pairs to nearest neighbors for training and inference.}
    \label{fig:D}
\end{figure}

\begin{table}
\centering
  \caption{Best estimates for test RMSEs obtained by Nearest Neighbor Twin Neural Network Regression (NNTNNR). Nearest neighbor training+inference beats nearest neighbor inference on 7 out of 9 data sets.
  }
 
  \begin{tabular}{l llrlr}\\

 & TNNR & NN inference & Gain & NN train+inference & Gain \\
       \cmidrule(r){2-2}\cmidrule(r){3-4}\cmidrule(r){5-6}

BC & 0.8234$\pm$0.0144 & 0.8162$\pm$0.0155 & 0.87\% & 0.8133$\pm$0.0152 & 1.23\% \\
BH & 3.3104$\pm$0.1202 & 3.2898$\pm$0.1202 & 0.62\% & 3.3563$\pm$0.1161 & -1.39\% \\
CS & 4.4731$\pm$0.1242 & 4.4458$\pm$0.1091 & 0.61\% & 4.4290$\pm$0.1174 & 0.99\% \\
EE & 0.7156$\pm$0.0204 & 0.7056$\pm$0.0193 & 1.4\% & 0.6825$\pm$0.0216 & 4.63\% \\
RCL & 0.0158$\pm$0.0002 & 0.0151$\pm$0.0003 & 3.98\% & 0.0140$\pm$0.0002 & 11.33\% \\
TF & 0.0050$\pm$0.0001 & 0.0028$\pm$0.0002 & 43.65\% & 0.0021$\pm$0.0003 & 57.26\% \\

WSB & 0.0233$\pm$0.0006 & 0.0224$\pm$0.0009 & 4.06\% & 0.0236$\pm$0.0009 & -1.01\% \\
WN & 0.6998$\pm$0.006 & 0.6951$\pm$0.0062 & 0.68\% & 0.6944$\pm$0.006 & 0.77\% \\
YH & 0.5977$\pm$0.0344 & 0.5184$\pm$0.0331 & 13.26\% & 0.5009$\pm$0.0333 & 16.19\%

  \end{tabular}
  \label{tab:nn_tnnr}
\end{table}
Let me first focus on NNTNNR with its two different training methods and the comparison to traditional TNNR. Nearest neighbor training is only trained on the $k$ nearest neighbors while training on all training pairs neglects this restriction. I emphasize that both versions of training obey the same principle for selecting the $k$ nearest neighbors during inference. 

These methods are compared in \fig{fig:D} where the baseline is set by standard TNNR (a discussion on the choice of  baseline can be found in \sec{sec:Neural Network Performance}:Neural Network Performance). At first glance, one can see that NNTNNR, trained on all training pairs, can outperform standard TNNR on all 9 data sets, assuming a suitable choice of the number of nearest neighbors $k$. While nearest neighbor training fails to beat standard TNNR on 2 data sets, it beats TNNR and NNTNNR trained on all pairs on 7 data sets. In these 7 data sets, there is a sweet spot where NNTNNR with nearest neighbor training outperforms at around 16 to 64 neighbors. In 3 of those data sets nearest neighbor training outperforms by a very large margin culminating in reducing the RMSE by $\approx 57\% $ on the TF data set, see \tab{tab:nn_tnnr}. Note that this is the data set with zero noise.

By having a look at the extreme choices of $k$ one can see interesting behaviour. When only using the very nearest neighbor as an anchor for inference $k=1$, one can see that for 7 out of 9 data sets both training versions underperform traditional TNNR. Further, in the limit of $k\rightarrow\infty$ one can confirm that the performance converges towards the performance of standard TNNR.

\subsection{NNTNNR vs k-NN}
\label{chapter:TNNR vs k-NN}
\begin{figure}
    \centering
    \includegraphics[width=0.95\textwidth]{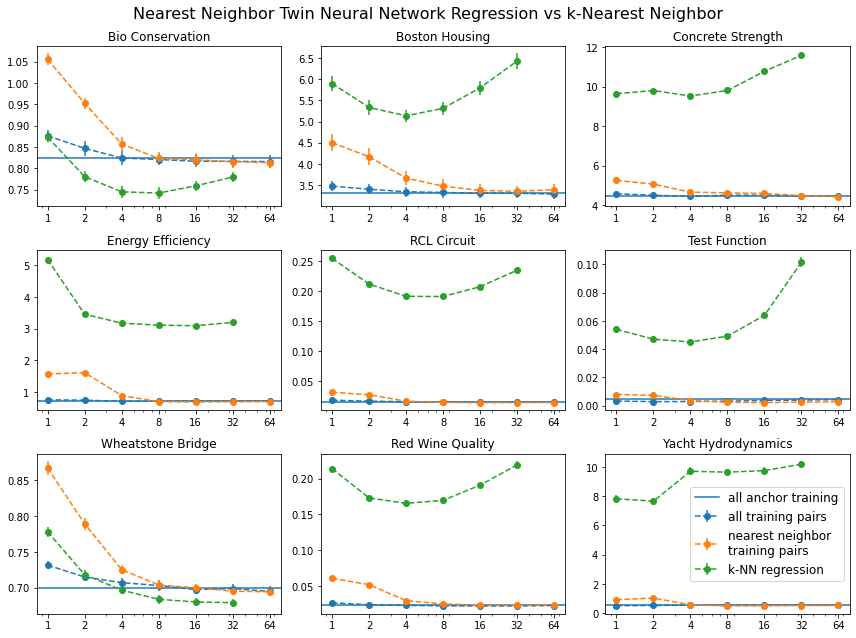}
    \caption{Comparison of k-NN regression and NNTNNR with different numbers of nearest neighbor training pairs measured by RMSE vs the number of neighbors for the dashed lines. The solid blue corresponds to normal TNNR with access to all possible pairs during training and inference. The blue dashed line restricts the pairs to be nearest neighbors during inference. The orange dashed line restricts the pairs to be nearest neighbors during training and inference. The green dashed line describes k-NN.}
    \label{fig:B}
\end{figure}
I visualize the behavior of k-NN and NNTNNR in \fig{fig:B} to display the very large performance differences between k-NN and NNTNNR. In this figure, one can clearly see, that NNTNNR beats k-NN regression by an enormous margin on 7 out of 9 data sets. However, there are two data sets, namely BC, WN where k-NN is the winner. This coincides with the data sets on which normal artificial neural network ensembles outperform TNNR as can be seen in the appendix in \fig{fig:A}. Further, the number of optimal TNNR neighbor anchors is much larger than the optimal number of neighbors in k-NN.

\section{Time Scaling}
\label{chapter:Time Scaling}

\begin{figure}
    \centering
    \includegraphics[width=0.95\textwidth]{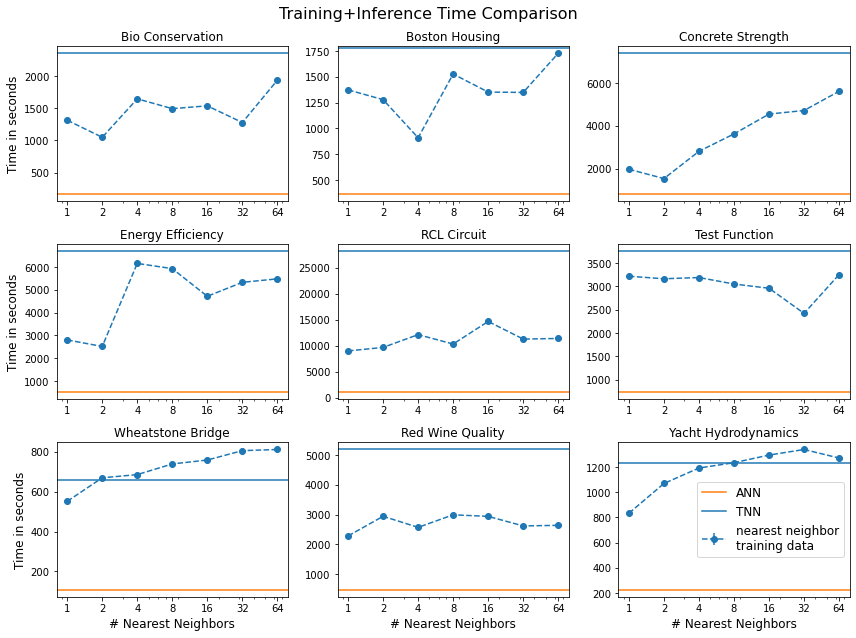}
    \caption{Training and inference time comparison in seconds between different versions of TNNR and ANN regression. The solid orange line indicates the time for training and inference in the case of ANNs. The solid blue line is the original TNNR. The blue dotted line indicates restricting the possible training (and inference) pairs to nearest neighbors, the x-axis corresponds to the number of neighbors. The error bars are omitted since they are larger than the plots.}
    \label{fig:C}
\end{figure}

In this section, I briefly discuss the time complexity of NNTNNR compared to k-NN regression and TNNR. Since k-NN does not involve any training procedure and in each inference step TNNR invokes a nearest neighbor search, the time required for NNTNNR is always greater than for k-NN regression. The comparison of NNTNNR with traditional TNNR is subject to an experimental analysis. As explored in \cite{wetzel2020twin}, the training time of twinned regression methods scales poorly towards larger data sets, mostly caused by the increase in the effective data set size through pairing of data points. While for many baseline algorithms a clear relationship between data set size and training time, for neural networks it is less known. As neural networks training time scales very favorably with training set size I focused on TNNR to test the training time improvement from only using nearest neighbor paring during training phase. In \fig{fig:C} one can see that there is a tendency for a reduced computational cost on most data sets. However, the training time scaling is too minor and inconsistent to use as a sole justification for using NNTNNR over traditional TNNR.

\section{Conclusion}
\label{chapter:conclusion}
In this article, I have developed a regression algorithm that combines k-nearest neighbor(k-NN) regression with twin neural network regression(TNNR), called nearest neighbor train neural network regression(NNTNNR). It is based on selecting the nearest neighbors of an unknown data point and averaging neural network-adjusted values belonging to these neighbors to arrive at the final prediction, as depicted in \fig{fig:NNTNNR}. 

This algorithm shows superior performance compared to both TNNR and k-NN, see \sec{chapter:results}:Results on most data sets. It is important to note that this extends to the outperformance over normal artificial neural networks, too, see \sec{sec:Neural Network Performance}:Neural Network Performance. 

During my analysis, I found that the performance behaviour of NNTNNR is much closer to TNNR than to k-NN regression. Further, there is a small improvement in performance if one only trains on training pairs containing similar pairings of neighbors as in the inference phase. This observation is likely based on the fact that differences between data points that are far away are more susceptible to a larger relative error.

NNTNNR might be very useful in solving regression problems where the data is sparse, gets outdated quickly, and shows different behavior in different parts of the training manifold. A prototypical example of such a data set is real estate in different cities, in a changing interest rate environment where sellers withhold their sales. Further, this algorithm might be useful in the solution of inverse problems.

\section{Acknowledgements}
I also acknowledge Compute Canada for computational resources. I thank the National Research Council of Canada for their partnership with Perimeter on the PIQuIL. Research at Perimeter Institute is supported in part by the Government of Canada through the Department of Innovation, Science and Economic Development Canada and by the Province of Ontario through the Ministry of Economic Development, Job Creation and Trade. This work was supported by Mitacs through the Mitacs Accelerate program.

\section{Declarations}
This work was supported by Mitacs and Homes Plus Magazine Inc. through the Mitacs Accelerate program. The code supporting this publication is available at \cite{githublink}. The data sets used in this work are described in \ref{sec:data}, they can either be found online at \cite{datasets} or generated from equations. I consent to publishing all code and data.
\\
\noindent *Conflicts of interest/Competing interests - not applicable\\
*Ethics approval - not applicable\\
*Consent to participate - not applicable\\
*Authors' contributions - not applicable

\newpage
\appendix
\onecolumn
\section{Data sets}
\label{sec:data}
\begin{table}[h!]
  \caption{Data sets
  }
  \centering
  \begin{tabular}{lllll}

  Name & Key & Size & Features & Type\\
\midrule
  Bio Concentration & BC &779&14&Discrete, Continuous\\
  Boston Housing & BH &506 & 13&Discrete, Continuous\\
  Concrete Strength & CS &1030&8&Continuous\\
  Energy Efficiency & EF &768&8& Discrete, Continuous \\
  RCL Circuit Current &RCL&4000&6&Continuous\\
  Test Function & TF & 1000 & 2 & Continuous\\
  Red Wine Quality & WN &1599&11&Discrete, Continuous\\
  Wheatstone Bridge Voltage &WSB&200&4&Continuous\\
  Yacht Hydrodynamics & YH &308&6&Discrete\\
  \end{tabular}
\label{tab:data}
\end{table}
The data sets used in this work are summarized in \tab{tab:data}, they can either be found online at \cite{datasets} or generated from the following equations. 

The test function (TF) data set created from the equation
\begin{align}
F(x_1,x_2)=x_1^3+x_1^2-x_1-1+x_1x_2+\sin(x_2)
\end{align}
and zero noise.

The output in the RCL circuit current data set (RCL) is the current through an RCL circuit, modeled by the equation
\begin{align}
I_0=V_0 \cos(\omega t)/\sqrt{R^2+(\omega L-1/(\omega C))^2}
\end{align}
with added Gaussian noise of mean 0 and standard deviation 0.1.

The output of the Wheatstone Bridge voltage (WSB) is the measured voltage given by the equation
\begin{align}
V=U(R_2/(R_1+R_2)-R_3/(R_2+R_3))
\end{align}
with added Gaussian noise of mean 0 and standard deviation 0.1.

\section{Neural Network Architectures}
\label{sec:nn_architecture}
Both the traditional neural network regression and twin neural network regression methods are build using the same architecture build using the tensorflow library \cite{Tensorflow2015}. They consist of two hidden layers with 128 neurons each and relu activation functions. The  final layer contains one single neuron without an activation function. I train the neural networks using the adadelta optimizer, and use learning rate and early stop callbacks that reduce the learning rate by 50\% or stop training if the loss stops decreasing. For this reason it is enough to set the number of epochs large enough such that the early stopping is always triggered, in all cases this is for a maximum of 2000 epochs for ANNs and 10000 epochs for TNNR. The batchsizes are in both cases 16.

\section{Neural Network Performance}
\label{sec:Neural Network Performance}

NNTNNR combines elements from k-NN regression and artificial neural networks. In order to evaluate the performance of NNTNNR we need to identify suitable baseline algorithms. While on the one hand, k-NN regression seems like an obvious choice, the type of neural network baseline is less clear. NNTNNR aggregates an ensemble of predictions of differences for each prediction corresponding to the underlying traditional regression problem. In this section, I briefly review the performance of ensembles of normal artificial neural network(ANN) regression and twin neural network regression for the purpose of finding an adequate baseline for the experiments surrounding NNTNNR. 

\begin{figure}
    \centering
    \includegraphics[width=0.95\textwidth]{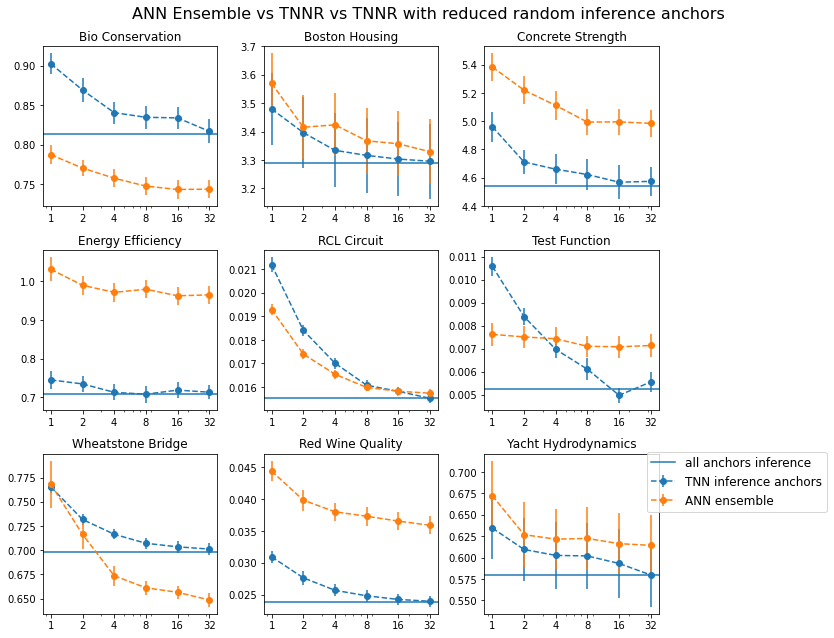}
    \caption{Comparison of different ensembles of ANNs and TNNR measured by RMSE. The blue solid line corresponds to training and evaluating one TNNR model on all possible training pairs and predicting the results with all possible anchors. The dashed blue line randomly chooses a subset of anchors during the inference phase. The dashed orange line indicates traditional ANN ensembles where multiple ANNs are trained independently.}    \label{fig:A}
\end{figure}

Let me start with discussing different kinds of ensembles and their effect on accuracy. \Fig{fig:A} contains the results of several experiments examining the performance of different ensemble types of ANN regression and TNNR. For each data set, the baseline results are the solid blue horizontal line, which represents the test RMSE after applying standard full anchor TNNR and the leftmost point of the orange dashed line which represents the results of applying a single ANN, confirming that TNNR almost always yields a lower RMSE than ANN regression. From there on the orange dashed line corresponds to creating an ensemble of ANNs by simply averaging over the predictions of similar ANNs trained using different starting initializations.

In an effort to close in on NNTNNR, during the inference phase, I apply TNNR with a reduced number of random anchor data points. The purpose is to mimic NNTNNR where the neighbors are chosen randomly instead of nearest neighbors. The results of this experiment is visualized in the blue dashed line. One can see that standard TNNR outperforms this random neighbor NNTNNR version (except for outliers within the errorbars). Further TNNR outperforms single ANNs and ensembles of ANNs on 7 out of 9 data sets. These are the same data sets where NNTNNR beats k-NN regression. Hence, the most suitable baseline for the experiments in this paper is standard TNNR.

\newpage
\phantom{a}


\end{document}